\documentclass[12pt, a4paper]{article}

\usepackage{abstract}
\usepackage{fancyhdr}
\usepackage{authblk}
\usepackage{graphicx}
\usepackage[square,sort,comma,numbers]{natbib}
\usepackage{amsfonts,amsmath,amssymb}
\usepackage{bm}
\usepackage{mathtools}

\usepackage{booktabs}
\usepackage{xcolor}
\usepackage{multirow}
\usepackage{rotating}
\usepackage{sistyle}

\SIthousandsep{,}

\DeclarePairedDelimiter{\abs}{\vert}{\vert}

\title{Trustworthy AI for Process Automation on a Chylla-Haase Polymerization Reactor}

\author[$\star$]{Daniel Hein}
\author[$\star\star$]{Daniel Labisch}
\affil[$\star$]{\small Siemens AG, Technology, Munich, Germany (e-mail: hein.daniel@siemens.com).\normalsize}
\affil[$\star\star$]{\small Siemens AG, Digital Industries, Karlsruhe, Germany (e-mail: daniel.labisch@siemens.com).\normalsize}

\newcommand{\abstractText}{\noindent
In this paper, genetic programming reinforcement learning (GPRL) is utilized to generate human-interpretable control policies for a Chylla-Haase polymerization reactor.
Such continuously stirred tank reactors (CSTRs) with jacket cooling are widely used in the chemical industry, in the production of fine chemicals, pigments, polymers, and medical products.
Despite appearing rather simple, controlling CSTRs in real-world applications is quite a challenging problem to tackle.
GPRL utilizes already existing data from the reactor and generates fully automatically a set of optimized simplistic control strategies, so-called policies, the domain expert can choose from.
Note that these policies are white-box models of low complexity, which makes them easy to validate and implement in the target control system, e.g., SIMATIC PCS 7.
However, despite its low complexity the automatically-generated policy yields a high performance in terms of reactor temperature control deviation, which we empirically evaluate on the original reactor template.
}

\usepackage{hyperref}
\hypersetup{colorlinks=true, urlcolor=blue, linkcolor=blue, citecolor=blue}

\begin{document}
	\bibliographystyle{plainnat}
		\maketitle
		\fancypagestyle{plain}{
			\lhead{}
			\fancyhead[R]{}
			\fancyhead[L]{\copyright 2021 the authors. 
				This is a preprint version. 
				The final version of this paper is available at \url{https://dl.acm.org/doi/10.1145/3449726.3463131} (DOI:10.1145/3449726.3463131).
				\newline
			\hrule
		}
			\renewcommand{\headrulewidth}{0pt}
			\fancyfoot[C]{\thepage}
		}
		\pagestyle{fancy}
		\fancyhead[R]{}
		\fancyhead[L]{}
		\renewcommand{\headrulewidth}{0pt}
		\fancyfoot[C]{\thepage}
		\setlength{\headheight}{53pt}
		
		\begin{abstract}
			\abstractText
			\newline
			\newline
		\end{abstract}
	
	\section{Introduction}

Utilizing AI-based methods in real-world industrial applications requires a certain amount of trust in the automatically-generated solution.
One way to build trust, is to have machine learning (ML) methods present the reasons for their decisions in a human-understandable way~\cite{doshi:17}.
On the one hand, one can attempt to interpret any useful system, even if it inherently is a black-box system, e.g., by explanation techniques for classifiers~\cite{ribeiro:16} or visualization methods for neural networks (NNs)~\cite{bach:15}.
On the other hand, interpretability can be evaluated via a quantifiable proxy, where at first, a model class, e.g., algebraic equations, is claimed to be interpretable and then algorithms are developed in order to optimize within this class~\cite{wang:16,maes:12,bischoff:13}.

\begin{figure}
	\centering
	\includegraphics[width=5in]{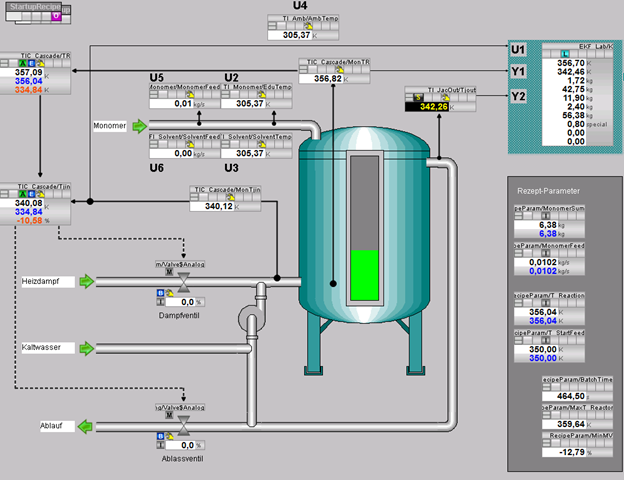}
	\caption[SIMATIC PCS 7 Chylla-Haase polymerization reactor template]{
		SIMATIC PCS 7 Chylla-Haase polymerization reactor template~\cite{siemens:18}
	}
	\label{figure:win_cc}
\end{figure}

For the latter, recently a new method called \textit{genetic programming reinforcement learning} (GPRL) has been proposed~\cite{hein:18c} and evaluated on challenging benchmark problems, like a hardware cart-pole demonstrator~\cite{hein:20} or a real-world inspired industrial benchmark~\cite{hein:19}.
In the present paper, we apply GPRL for the first time on a SIMATIC PCS~7 Chylla-Haase polymerization reactor template (see Fig.~\ref{figure:win_cc})~\cite{siemens:18}, to improve the process automation controller while keeping the policy solution interpretable and thus trustworthy.
The template exemplifies the conventional control strategy using a realistic but simple simulation model directly built in SIMATIC PCS~7. 
It aims as a template for real applications.

If a process plant runs safely and reliably during operation, the operator always strives to maximize the economic yield. 
Existing degrees of freedom are used to design the operation as optimally as possible without violating existing limitations and boundary conditions.
This can be done through manual intervention by the operator based on experience, or through the use of various optimization methods.

Optimization methods can basically be divided into two classes, simulation-based and data-based tools, although mixed forms are also possible. 
Strictly speaking, the optimization works with a simulation model in the first case and with a data-based model in the second case. 
An optimization based on a \textit{general PROcess Modelling System} (gPROMS) is an example for the first case.
Most data-based approaches create black-box results like neural networks.
These models can show impressive results, but most of them are too complex to be interpreted or validated.

Hence, available optimization methods for process industries are either interpretable expert-generated process controllers or non-interpretable AI-generated process controllers. 
Herein, we show how the ML method genetic programming can be utilized together with supervised AI-based regression models to automatically generate interpretable process controllers. 
By doing so, the manual effort to generate interpretable process controllers can be reduced significantly.
Furthermore, novel unknown control strategies can be found, which improve the quality key performance indicators (KPIs) of the controller.

In contrast to simulation-based approaches, this method significantly reduces the modeling efforts. 
Manually creating a simulation model is very costly in general. 
Even if a model already exists, e.g., from the planning phase, the adaption and parametrization of the model to represent the real behavior is still very time-consuming. 
Furthermore, many simulation models from the planning phase are just static ones which couldn’t be easily extended to a dynamic version. 
The data-based approach GPRL does not require all these efforts. 

Furthermore, generating black-box controllers by ML can be prone to overfit to some process model inaccuracies. 
Black-box controllers normally have hundreds of parameters which are automatically tuned to yield a high performance with respect to the predicted control quality. 
However, in many cases model inaccuracies and loopholes can be exploited by the ML process which result in controllers that only performing well on the imperfect data-based models but not on the real process. 
Interpretable controllers by genetic programming, formulated using the established controller building blocks are more robust to process model inaccuracies. 
Furthermore, experts of the respective process are able to identify contradictions and implausibilities in interpretable controllers.

The resulting process controllers can either directly replace existing control structures or can also be used as a superimposed optimization strategy.
This flexibility combined with the interpretable controller function enables the process owner to balance the conflicting goals of optimizing performance and guaranteeing safety and reliability of the process.

In Sections~\ref{section:chylla-haase} and \ref{section:gprl}, the interface of the reactor template and the GPRL method are introduced, respectively.
In Section~\ref{section:interpretable_process_control}, we show how GPRL uses available simulation data from a Chylla-Haase reactor to generate interpretable policies.
In the evaluation phase of the experiment, we show that the newly found policies reduce the control error by up to 37~\%.
	
	\section{Chylla-Haase polymerization reactor}
\label{section:chylla-haase}

Continuously stirred tank reactors (CSTRs) with jacket cooling are widely used in the chemical industry, in the production of fine chemicals, pigments, polymers, and medical products.
CSTRs can be run in batch, semi-batch or continuous operation.~\cite{siemens:18}

To perform an accurate control, online optimization and monitoring of processes, precise knowledge of the status of the respective process is required at all times. 
However, only in rare cases it is possible to measure all the states of the reactor using real sensor equipment.
Often, some process variables can not be determined at all or only with great technical effort.

Simulating the dynamic process model in a soft sensor in parallel to the real process, allows the automation system to observe all modeled inner states of the reactor even if they cannot be registered by measurement equipment.
The Chylla-Haase reactor, considered in our experiments, contains a universal stirred tank reactor that is employed for the production of diverse polymers with different recipes. 
Here, an extended Kalman filter serves as a soft sensor for monitoring the chemical reaction.

The values calculated online by the soft sensor can be used in various ways~\cite{siemens:18}:
\begin{itemize}
	\item Calculation of the current speed of reaction and the heat released by the exothermic reaction,
	\item Calculation of the monomer mass remaining in the reactor, and
	\item Calculation of heat transfer from reactor to cooling jacket in order to detect the build-up of deposition.
\end{itemize}

A frequently considered control problem statement for Chylla-Haase reactors, is to maintain reaction temperature of a semi-batch polymerization reactor at a setpoint in spite of a variety of conditions ranging from raw material impurity, product variety, repeated batching and varying ambient temperature. 
Although the reactor configuration and control requirements appear rather simple, in real-world applications it is quite a challenging problem to tackle.~\cite{bhat:98}

The rector template (SIMATIC PCS 7 project and documentation) we used in our experiments is available online\footnote{\url{https://support.industry.siemens.com/cs/de/de/view/109756215}}.
The production of polymer is maintained by an \textit{sequential function chart} (SFC) recipe, where the simulation of one batch takes between 10 to 40 minutes, depending on the process parameters.
The operator screen as user interface of the template is depicted in Fig.~\ref{figure:win_cc}.

The reactor template can be influenced by two variables either given by the operator or using a superimposed optimization strategy during the batch:
\begin{itemize}
	\item Reactor temperature setpoint $\hat{T}$, and
	\item Monomer feed setpoint $\hat{M}$.
\end{itemize}
These variables are called decision variables or actions.

The task is to process a certain amount of monomer into polymer, given a certain reactor setpoint.
According to the recipe provided, the reactor is heated up to a certain start temperature.
If this temperature is reached, the monomer is added to the stirred tank reactor at a certain monomer feed rate $\hat{M}$.
As soon as a pre-defined amount of monomer is processed , the batch is finished and the tank gets drained.
Note that the reactor temperature has to be controlled constantly by heating and cooling, since the reaction temperature is highly dependent on the amount of polymer inside the tank.

The default method to control the temperature is to use a proportional– integral–derivative (PID) controller.
Today PID control is still one of the most common control strategies, since it is very often applied at the lowest level of the control hierarchy in industrial systems~\cite{aastrom:04}. 
In 2002, \citeauthor{desborough:02} conducted a survey of more than \num{11000} controllers in the refining, chemicals, and pulp and paper industries which revealed that 97~\% of regulatory controllers had a PID structure~\cite{desborough:02}.
However, PID controllers can have severe drawbacks. 
They use only limited process information and the design criterion sometimes yields closed loop systems with poor robustness~\cite{aastrom:01}.

Hence, it is expected that an RL-based controller, which is potentially non-linear and can utilize all available process variables, can outperform standard PID regulation on many real-world use cases.
The action and state variables of the reactor template, which are available for an RL controller, are given in Table~\ref{table:variables}.
The action variables in $\bm{a}$ are the decision variables of the reactor template, whereas the state variables in $\bm{s}$ contain the intended setpoint $S$ as well as sensor measurements from the reactor.
Note that $S$ is not a variable modeled by the reactor simulation, but it represents the true intended temperature setpoint of the operator.
The default controller sets $\hat{T}=S$, i.e., the operator's intended setpoint is directly used as input for the underlying PID regulator.

\begin{table}
\begin{center}
\begin{tabular}{cllll}  
	\toprule
	& Variable & Description & Unit & Range\\
    \midrule
    \multirow{2}{*}{\begin{turn}{90}\rule[2pt]{5pt}{0.5pt} $\bm{a}$ \rule[2pt]{5pt}{0.5pt}\end{turn}} & $\hat{T}$ & Reactor temperature setpoint & K & 352-365\\
    & $\hat{M}$ & Monomer feed setpoint & kg/s & 0.005-0.015\\
    \midrule
    \multirow{2}{*}{\begin{turn}{90}\rule[2pt]{28pt}{0.5pt} $\bm{s}$ \rule[2pt]{28pt}{0.5pt}\end{turn}} & $S$ & Intended setpoint & K & 352-365\\
    & $T$ & Reactor temperature & K & -\\
    & $M$ & Monomer mass & kg & -\\
    & $P$ & Polymer mass & kg & -\\
    & $UA$ & Heat loss coefficient & kW/K & 0-1\\
    & $Q$ & Reaction heat flow & kW & -\\
    \bottomrule	
\end{tabular}
\caption[Reactor variables]{
	Reactor variables of action $\bm{a}$ and state $\bm{s}$
	}
\label{table:variables}
\end{center}
\end{table}
	
	\section{Genetic programming reinforcement learning}
\label{section:gprl}

The GPRL approach learns policy representations which are basic algebraic equations of low complexity~\cite{hein:18c}. 
Given that GPRL can find rather short (non-complex) equations, it is expected to reveal substantial knowledge about underlying coherencies between available state variables and well-performing control policies for a certain RL problem.
By the term complexity of a policy, we refer to its human-interpretable form, e.g., a simple algebraic equation can be easily understood by a domain expert, hence it is of low complexity, whereas a non-linear system of equations, like a neural network, cannot be validated by a human expert, yielding a high complexity. 

Note that GPRL is not only optimizing a certain trade-off between policy complexity and performance, but rather it evolves a whole optimized Pareto front from which the domain expert can choose a trustworthy and well-performing solution.

\subsection{Population-based reinforcement learning}

Generally, reinforcement learning (RL) is distinguished from other computational approaches by its emphasis on an agent learning from direct interaction with its environment.
This approach, which does not rely on exemplary supervision or complete models of the environment, is referred to as \textit{online learning}.
However, for many real-world problems online learning is prohibited for safety reasons.
For example, it is not advisable to deploy an online RL agent, who starts by applying an arbitrary initial policy, which is subsequently improved by exploitation and exploration, on safety-critical domains like process industries or power plants.
For this reason, \textit{offline learning} is a more suitable approach for applying RL methods on already collected training data and system models, to yield RL policies.

Offline RL is often referred to as \textit{batch learning} because it is based solely on a previously generated batch of transition samples from the environment.
The batch data set $\mathcal{D}$ contains transition tuples of the form $(\bm{s}_t,\bm{a}_t,\bm{s}_{t+1},r_{t+1})$, where the application of action $\bm{a}_t$ in state $\bm{s}_t$ resulted in a transition to state $\bm{s}_{t+1}$ and yielded a reward $r_{t+1}$, where $t$ denotes a discrete time step.

Herein, we use \textit{model-based value estimation} to estimate the performance of a policy.
In the first step, supervised ML is applied to learn approximate models of the underlying environment from transition tuples $(\bm{s}_t,\bm{a}_t,\bm{s}_{t+1},r_{t+1})$ as follows:
\begin{equation}
	\tilde{g}(\bm{s}_t,\bm{a}_t)\leftarrow\bm{s}_{t+1},\quad \tilde{r}(\bm{s}_t,\bm{a}_t,\bm{s}_{t+1})\leftarrow r_{t+1}.
	\label{eq:model}
\end{equation}
Using models $\tilde{g}$ and $\tilde{r}$, the value for policy $\pi$ of each state $\bm{s}$ in the data batch can be estimated by computing the value function $\tilde{v}_\pi(\bm{s})$.
Hence, using model-based value estimation means performing trajectory rollouts on system models for different starting states:
\begin{align}  
	\tilde{v}_{\pi}(\bm{s}_t) &= \sum_{k=0}^{\infty}\gamma^k \tilde{r}(\bm{s}_{t+k},\bm{a}_{t+k},\bm{s}_{t+k+1}),\label{eq:model_based_evaluation} \\
	\textnormal{with}\quad \bm{s}_{t+k+1} & = \tilde{g}(\bm{s}_{t+k},\bm{a}_{t+k}),\quad \bm{a}_{t+k}=\pi(\bm{s}_{t+k}),
\end{align}
and discount factor $0\leq\gamma\leq1$.

Since we are searching for interpretable solutions, the resulting policies have to be represented in an explicit form.
\textit{Policy search} yields explicit policies which can be enforced to be of an interpretable form.
Furthermore, policy search is inherently well-suited for being used with population-based optimization techniques, like genetic programming (GP). 

The goal of using policy search for learning interpretable RL policies is to find the best policy among a set of policies that is spanned by a parameter vector $\bm{x}\in \mathcal X$, where $\mathcal X$ is the space of all interpretable policy parameters for this RL task. 
Herein, a policy corresponding to a particular parameter value $\bm{x}$ is denoted by $\pi[\bm{x}]$.
The policy's performance, when starting from $\bm{s}_t$ is measured by its value function given in Eq.~\eqref{eq:model_based_evaluation}.
Furthermore, including only a finite number of $T>1$ future rewards for the model-based value estimation from Eq.~\eqref{eq:model_based_evaluation} yields
\begin{equation}
	\begin{aligned}    
		\tilde{v}_{\pi[\bm{x}]}(\bm{s}_t) &= \sum_{k=0}^{T-1}\gamma^k \tilde{r}(\bm{s}_{t+k},\bm{a}_{t+k},\bm{s}_{t+k+1}), \\
		\textnormal{with}\quad \bm{s}_{t+k+1} & = \tilde{g}(\bm{s}_{t+k},\bm{a}_{t+k}) \quad \textnormal{and}\quad \bm{a}_{t+k}=\pi[\bm{x}](\bm{s}_{t+k}).
	\end{aligned}
\label{eq:model_based_value}
\end{equation}

To rate the performance of policy ${\pi[\bm{x}]}$, the value function is used as follows:
\begin{equation}
	\overline{\mathcal{R}}_{\pi[\bm{x}]}=\frac{1}{\abs{\mathcal{S}}}\sum_{\bm{s}\in\mathcal{S}}\tilde{v}_{\pi[\bm{x}]}(\bm{s}),
\label{eq:return}
\end{equation}
with $\overline{\mathcal{R}}_{\pi[\bm{x}]}$ being the average discounted return of ${\pi[\bm{x}]}$ on a representative set of test states $\mathcal{S}$.

\begin{figure}
	\begin{center}
		\includegraphics[width=3.4in]{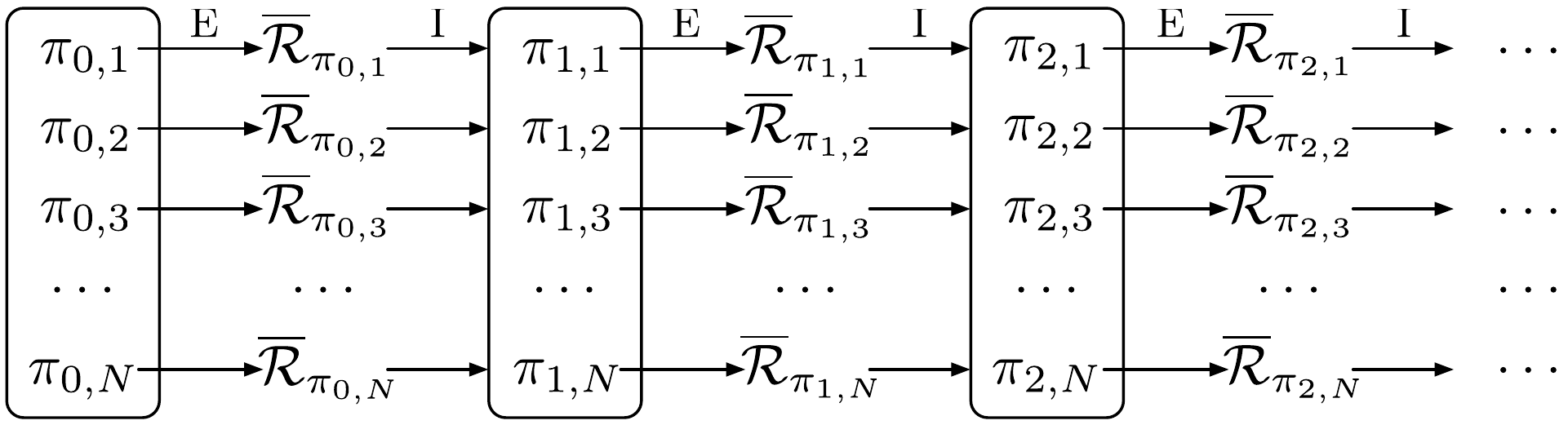}
		\caption{Population-based policy search iterating policy evaluation (E) and policy 	improvement (I)}
		\label{fig:policy_search_population}
\end{center}
\end{figure}

In population-based RL, the interest lies in populations of policies, which means that multiple different policies exist at the same time in one policy iteration step.
Fig.~\ref{fig:policy_search_population} depicts how the return values of all population members drive the improvement step (I) of the whole population.
The evaluation step (E) is performed individually for every member.
Applying population-based evolutionary methods to policy search has already been successfully achieved in the past~\citep{chin:98,gomez:06,chang:07,wilson:18}.

\subsection{Genetic programming}

GP has been utilized for creating system controllers since its introduction~\citep{koza:92}. 
Since then, the field of GP has grown significantly and has produced numerous results that can compete with human-produced solutions~\citep{koza:10}, including system controllers~\citep{keane:02,shimooka:98}, game playing~\citep{gearhart:03,wilson:18}, and robotics~\citep{downing:01,kamio:05}. 

GP encodes computer programs as sets of genes and then modifies (evolves) them using a so-called genetic algorithm (GA) to drive the optimization of the population, by applying selection and reproduction to the population.
The basis for both concepts is a fitness value, which represents the quality of performing the predefined task for each individual.
Selection means, that only the best portion of the current generation will survive each iteration and continue existing in the next generation.
Analogous to biological sexual breeding, two individuals are selected for reproduction based on their fitness, and two offspring individuals are created by crossing their chromosomes.
Technically, this is realized by selecting compatible cutting points in the function trees and subsequently interchanging the subtrees beneath these cuts.
The two resulting individuals are introduced to the population of the next generation.
Herein, we applied tournament selection~\citep{blickle:95} for selecting the individuals to be crossed.

To rate the quality of each policy candidate, a fitness value has to be provided in order for the GP algorithm to advance.
For GPRL, the fitness of each individual is calculated by generating trajectories using the model-based return estimation from Eq.~\eqref{eq:return}.

The overall GA used in the experiments is given as follows:
\begin{enumerate}
	\item Randomly \textbf{initialize} the population of size $N$
	\item \textbf{Determine fitness} value of each individual using Eq.~\eqref{eq:return} (in parallel)
	\item \textbf{Evolve} next generation
	\begin{enumerate}
		\item \textbf{Crossover} (depending on crossover ratio $r_\text{c}$)
		\begin{enumerate}
			\item Select individuals by \textbf{tournament selection}
			\item Cross two tournament winners
			\item Add resulting individuals to new population
		\end{enumerate}
		\item \textbf{Reproduction} (depending on reproduction ratio $r_\text{r}$)
		\begin{enumerate}
			\item Select individuals by \textbf{tournament selection}
			\item Add tournament winner to new population
		\end{enumerate}
		\item Automatic \textbf{cancelation} and terminal \textbf{mutation} (depending on auto cancel ratio $r_\text{a}$ and terminal mutation ration $r_\text{m}$)
		\begin{enumerate}
			\item Apply automatic \textbf{cancelation} on all individuals
			\item Add canceled individuals according to $r_\text{a}$
			\item Select best individuals for each complexity level of old population
			\item Randomly \textbf{mutate} float terminals using normal distribution $\mathcal{N}$: $z'\sim z+0.1z\cdot\mathcal{N}(0,1)$, where $z$ and $z'$ are the original and the mutated float terminals, respectively; and create $N\cdot r_\text{a}$ adjusted individuals from each best
			\item Determine fitness value of each individual (in parallel)
			\item Add best adjusted individuals to new population according to $r_\text{m}$
		\end{enumerate}
		\item \textbf{Fill} new population with new randomly generated individuals (new individuals ratio  $r_\text{n}$)
		\item \textbf{Determine fitness} value of each individual using Eq.~\eqref{eq:return} (in parallel)
		\item If none of the stopping criteria is met
		\begin{enumerate}
			\item Go back to 3.
		\end{enumerate}
	\end{enumerate}
	\item \textbf{Return} best individual found so far for each complexity level
\end{enumerate}

For the experiments described in Section~\ref{section:interpretable_process_control}, the following new population ratios have been used: $r_\text{c}=0.45$, $r_\text{r}=0.05$, $r_\text{m}=0.1$, $r_\text{a}=0.1$, $r_\text{n}=0.3$.
Stopping criteria can be a maximum number of iterations, a certain algorithm runtime, a pre-defined performance value, etc.
In our experiments, we stopped the GA after 100 iterations.

GPRL has been implemented using the open source evolutionary computation framework DEAP\footnote{\url{https://github.com/DEAP/deap}} which provides an excellent and widely-used Python implementation of genetic programing using prefix trees~\cite{fortin:12}.
A detailed explaination of the GA utilized in GPRL, including a description of automatic cancelation and parameter motivation, can be found in~\cite{hein:19}.
	
	\section{Interpretable process controllers for a Chylla-Haase polymerization reactor}
\label{section:interpretable_process_control}

\begin{figure}
	\centering
	\includegraphics[width=3.3in]{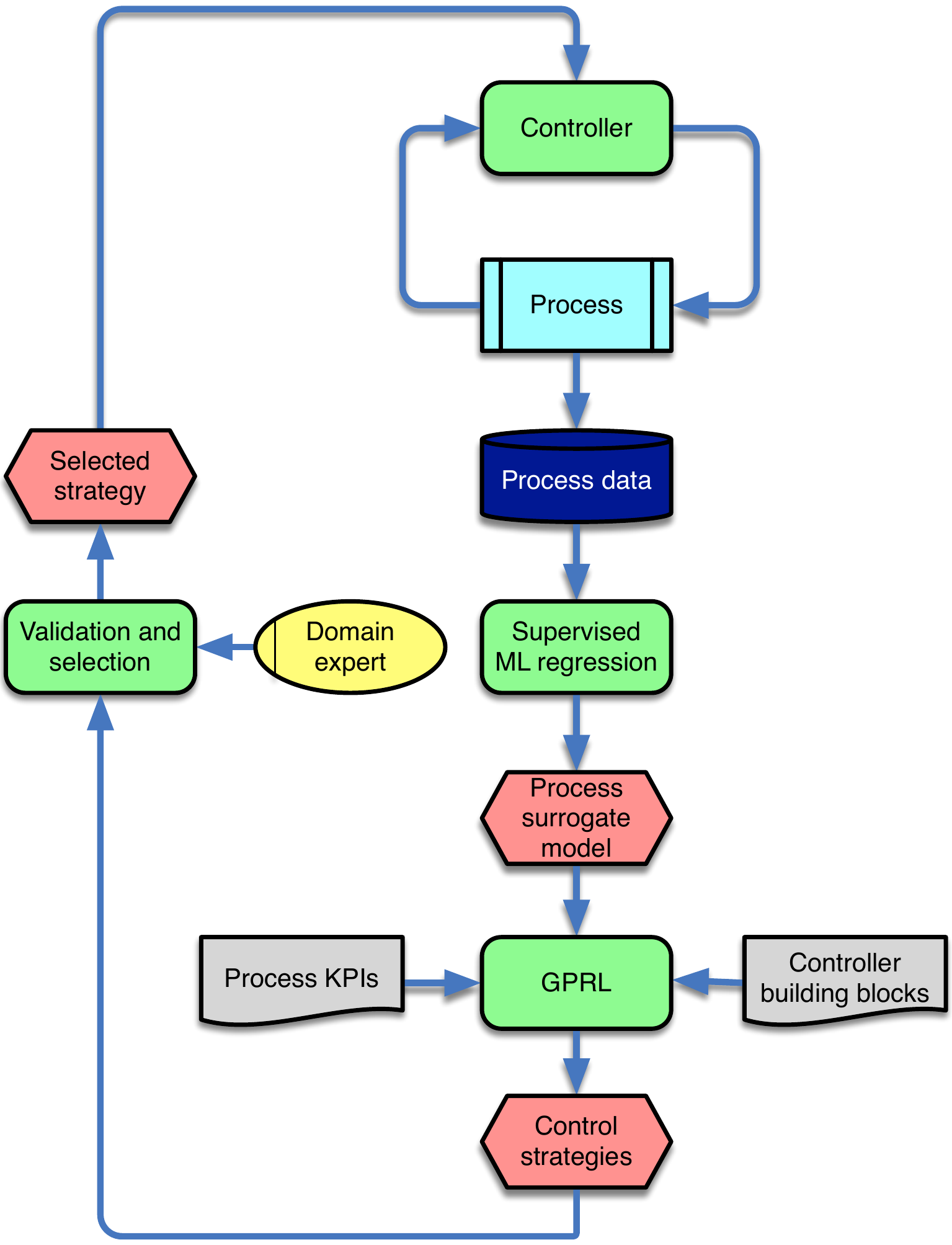}
	\caption[EPC of generating interpretable process controllers]{
		EPC of generating interpretable process controllers using GPRL. 
		Note that the Domain expert is integrated into the optimization loop. 
		However, generating interpretable control strategies is fully automatized.
	}
	\label{figure:overview}
\end{figure}

The framework of learning interpretable process controllers contains two ML steps.
Firstly, available data from the process is used to generate a process model.
Here, we used supervised ML to train weights of a recurrent neural network as system identification.
Secondly, GPRL is performed on this recurrent model to produce several Pareto-optimized process controllers.
The controllers are optimized towards the process KPIs utilizing the platform-specific process control building blocks.
An event-driven process chain (EPC) flowchart of applying GPRL for process optimization is depicted in Fig.~\ref{figure:overview}.

\subsection{Recurrent system identification}

To generate a batch data set $\mathcal{D}$ with sufficient experience, 100 different exploration process recipes have been used.
In the classic control strategy, using a PID temperature controller, the controller receives a constant setpoint. 
As the controller needs to handle both, heating and cooling during the production, the performance is not optimal and significant control deviations could be observed. 
This will be shown in Section~\ref{section:evaluation} together with the results. 
For safety and acceptance reasons, GPRL is not intended to replace the PID controller but to be superimposed on it. 
Therefore, the setpoint temperature $\hat{T}$ of the PID controller becomes the action of GPRL and a new overall setpoint $S$ is introduced (see Table~{\ref{table:variables}). 
Modifying $\hat{T}$ reveals new degrees of freedom not used before in the operation of the reactor.
Therefore, new training data needs to be created wherein $\hat{T}$ is changed.
For our experiments, the setpoint has been changed once per recipe between 352 and 365 K at a random batch runtime between 100 and 600 s.
In process optimization, such step attempts are a common method of safely exploring in a bounded subspace of the control problem.

The response of the process on these changes is then recorded and exported to the data batch.
In the next step, the data is subsampled on a 10~s grid and then normalized, i.e., average 0 and standard deviation 1.

The system model $\tilde{g}$ is a recurrent neural network (RNN) with the task to predict the values of $\bm{s}_{t+1}=(T_{t+1}, M_{t+1}, P_{t+1}, UA_{t+1}, Q_{t+1})$, $\bm{s}_{t+2},\ldots,\bm{s}_{t+F}$ from the inputs $\bm{s}_{t}=(T_{t}, M_{t}, P_{t}, UA_{t}, Q_{t}),\bm{s}_{t-1},\ldots$, $\bm{s}_{t-H+1}$ and $\bm{a}_{t}=(\hat{T}_t, \hat{M}_t),$ $\bm{a}_{t-1},\ldots,\bm{a}_{t-H+1}$.
The RNN is unfolded $H=10$ time steps into the past and $F=10$ time steps into the future. 
In each time step, the observable variables of the past and present are inputs. 
Whereas, in the future branch of the RNN only the control variables $\hat{T}$ and $\hat{M}$ are used as input. 
The topology of the RNN is described in~\cite{schaefer:05} as \textit{Markov decision process extraction network}.
The model was implemented as a tensorflow graph using RNN cells with two hidden layers and 20 $\tanh$ activation neurons on each layer.

Note that an $F=10$ overshooting is used, which means that the regression error used to optimize the network weights, is not only computed on the next time step, but as an average error over the next 10 time steps.
In our experience, this generally helps to yield more robust predictions for models which are used in closed loop simulations, where predictions have to be made on the models own, possibly inaccurate, previous predictions.

After \num{10000} training episodes, the quadratic loss of the prediction has fallen to 0.0017 on the trainings set, 0.0022 on the validations set, and 0.0021 on the generalization set (see Fig.~\ref{figure:model_errors_learning}).
The sets contained 70, 20, and 10 step attempt time series for training, validation, and generalization, respectively.
Fig.~\ref{figure:model_errors_prediction} depicts the average absolute loss for the predicted reactor temperature $T$ over the 10 overshooting time steps.
It is shown, how the prediction quality is decreasing for more distant time steps.
However, a validation error increase from 0.019 to 0.026 on a prediction horizon of 100 seconds (10 time steps) still yields an adequate system model.

\begin{figure}
	\centering
	\includegraphics[width=3.3in]{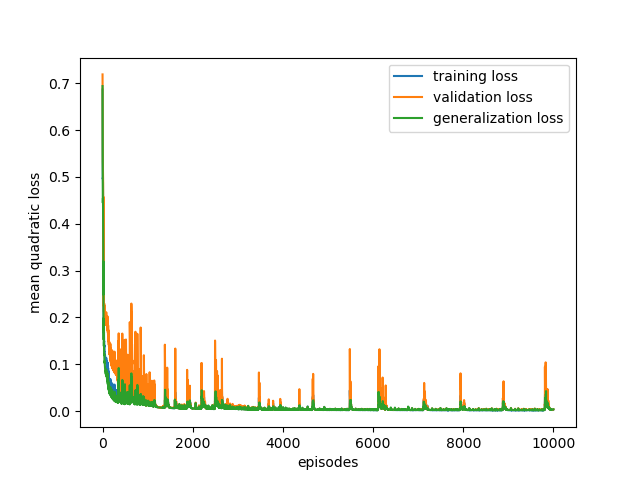}
	\caption[Learning curve]{
		Learning curve of the RNN surrogate model training
	}
	\label{figure:model_errors_learning}
\end{figure}

\begin{figure}
	\centering
	\includegraphics[width=3.3in]{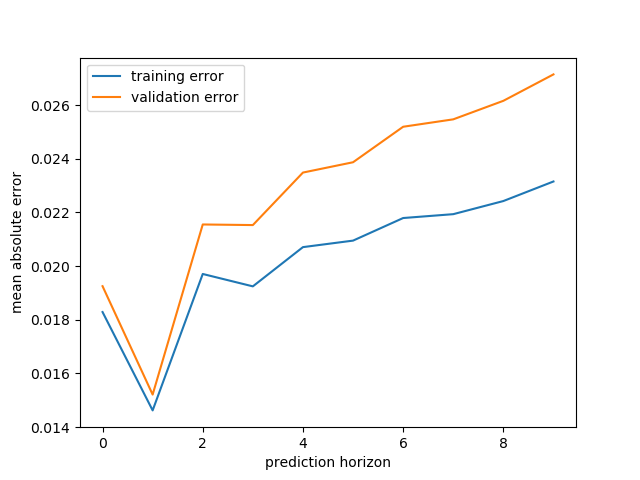}
	\caption[Prediction error]{
		Prediction error of the trained model RNN surrogate model over ten future time steps
	}
	\label{figure:model_errors_prediction}
\end{figure}

\subsection{Learning an interpretable process controller}

The RNN, trained in the previous step, can now be used as model $\tilde{g}$ together with the reward function $\tilde{r}$ to rate the performance of policy candidates.
In case of the temperature control task we consider here, the reward is computed by $\tilde{r}(\bm{s}_t)=r_t=-(S-T_t)^2$, where $S_t$ is the desired setpoint and $T_t$ is the predicted reactor temperature at time step $t$.
According to Eq.~\eqref{eq:model_based_value}, the controller parameters $\bm{x}$ are tested on 100 starting states $\mathcal{S}$ drawn from the data batch $\mathcal{D}$.
The trajectories produced by closed loop evaluation of policy $\pi$ with parameters $\bm{x}$ and system model $\tilde{g}$, have a length of $T=10$.
Hence, the performance of $\bm{x}$ can be estimated by Eq.~\eqref{eq:return}.

The building blocks of the genetic programming algorithm are basic algebraic functions $\{+,-,*,/\}$, state and action variables $\{\bm{s}, \bm{a}\}$, and dead time blocks storing past values of the variables up to 50 seconds $\{\square_{t},\square_{t-10},\ldots,$ $\square_{t-50}\}$, with $\square$ an arbitrary variable from $\bm{s}$ or $\bm{a}$.
Here, we are searching for a policy that serves as a setpoint process controller which dynamically changes setpoint $\hat{T}$ to minimize the distance between the desired setpoint $S$ and the actual reactor temperature $T$.
The monomer setpoint $\hat{M}$ has been fixed to the value 0.015.
Note that the policy can incorporate state and action values from the past and is evaluated in a closed loop manor for ten time steps during training.
Consequently, it will not greedily try to reach the desired setpoint in the next step, but in the long run try to avoid overshooting and maximize the reward instead.

After 100 iterations of GPRL with 500 individuals in each generation, the performance of the best policies found are visualized in Fig.~\ref{figure:pareto_front}.
It is easy to see that individuals of lower complexity have a higher penalty (lower reward) compared to individuals of higher complexity.
However, the estimated performance seems to not improve significantly for individuals of complexity 10 or higher.
After discussion with domain experts, an individual of complexity 9 and an estimated penalty of around 392 has been selected (highlighted in red).
The selected individual can be represented as the following equation:
\begin{equation}
	\hat{T}_t=T_{t-30}-2T_t+2S-1.
	\label{eq:policy}
\end{equation}

\begin{figure}
	\centering
	\includegraphics[width=3.3in]{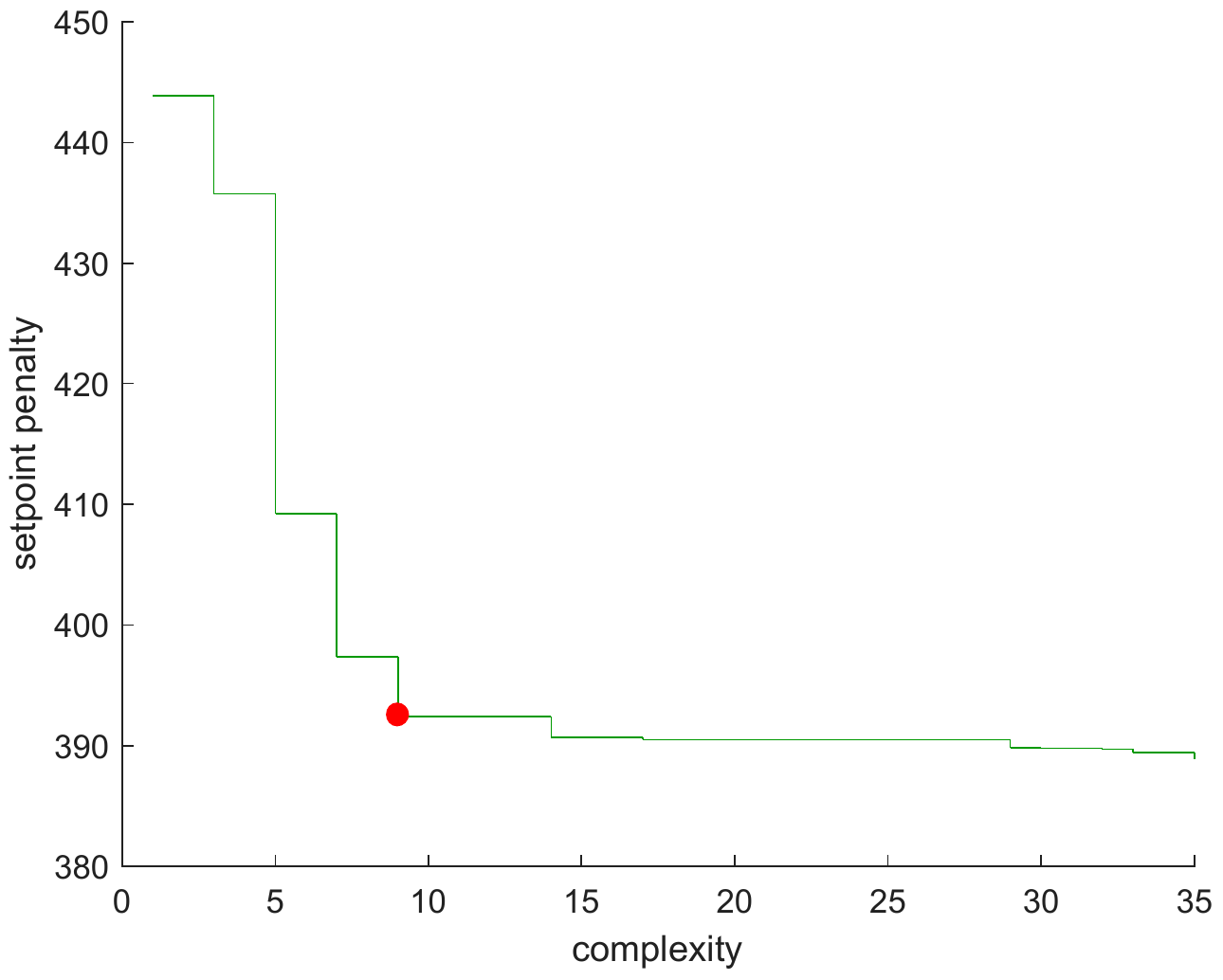}
	\caption[Pareto front]{
		Pareto front generated by GPRL. 
		According to the domain expert, the selected policy (red dot) is a good compromise between setpoint penalty (392) and complexity (9).
	}
	\label{figure:pareto_front}
\end{figure}

This equation can easily be interpreted thanks to its simple structure.
The reactor temperature setpoint is dynamically changed with respect to the reactor temperature 30~s ago, the current temperature, and the intended setpoint.
Note that all variables ($T$, $\hat{T}$, and $S$) in the equation have been normalized by average 359.12 and standard deviation 6.47.

\subsection{SIMATIC PCS 7 integration}

The policy found by GPRL and represented by Eq.~\eqref{eq:policy}, can easily be implemented in SIMATIC PCS 7, since GPRL utilized only components from the pre-defined building blocks which are available in the target system.
Fig.~\ref{figure:scl} shows a screenshot of the respective \textit{structured control language} (SCL) code.
Note, that since this is the native language of the system at hand, the new policy is easy to integrate and fast in execution.
Domain experts are able to fully understand the code and can easily make adjustments.
All of this would be impossible if we were to use a black-box policy.

\begin{figure}
	\centering
	\includegraphics[width=5in]{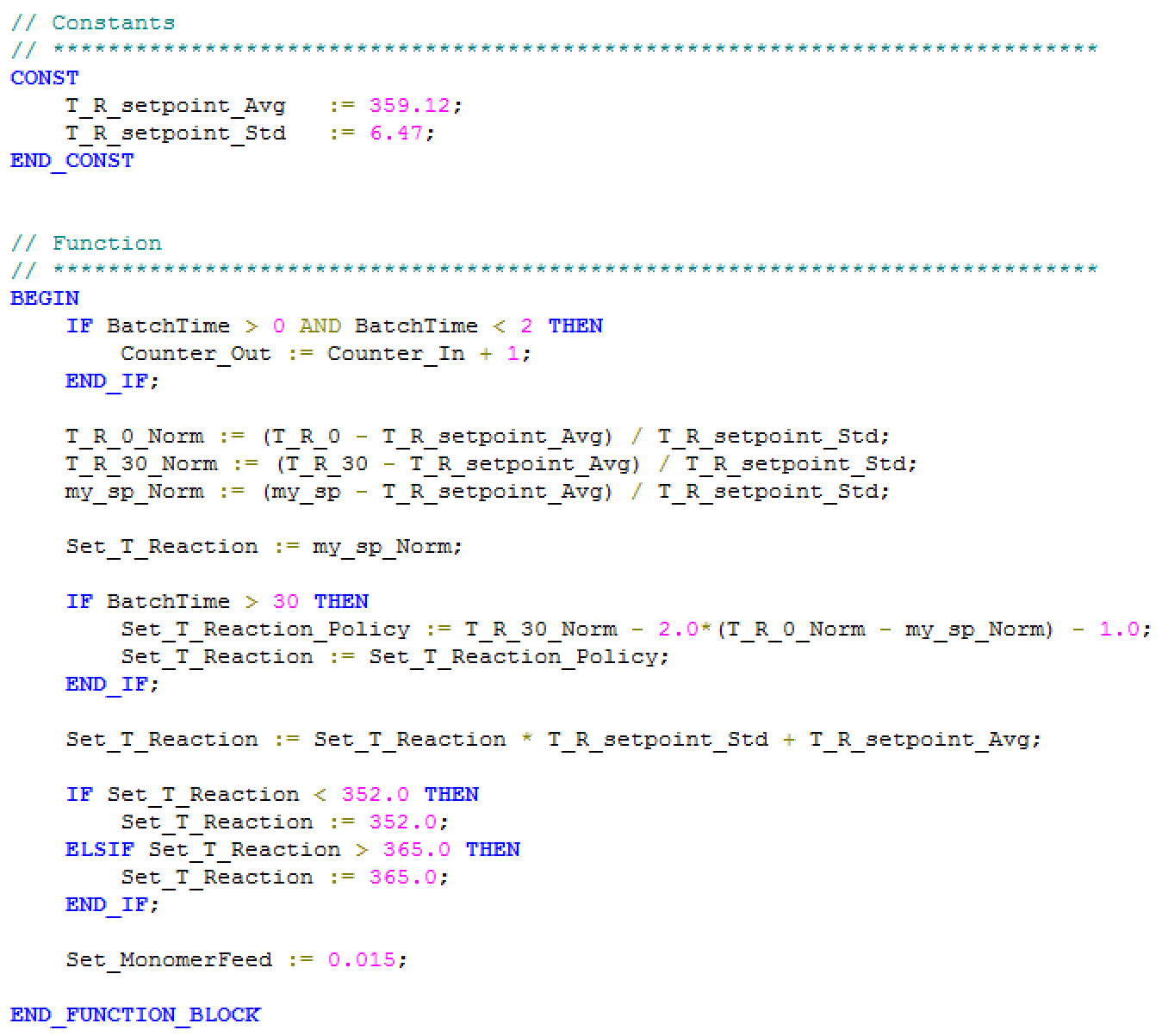}
	\caption[SCL]{
		SCL script with the implemented GPRL control strategy
	}
	\label{figure:scl}
\end{figure}

After generating the SCL code, the new control strategy can be loaded into a standard \textit{continuous function chart} (CFC) building block.
Fig.~\ref{figure:cfc} shows the CFC diagram with our new policy block (FB10) and the additional dead time block (DeadTime) with 30~s below.
The policy is now fully integrated into the automation system and can be evaluated.

\begin{figure}
	\centering
	\includegraphics[width=5in]{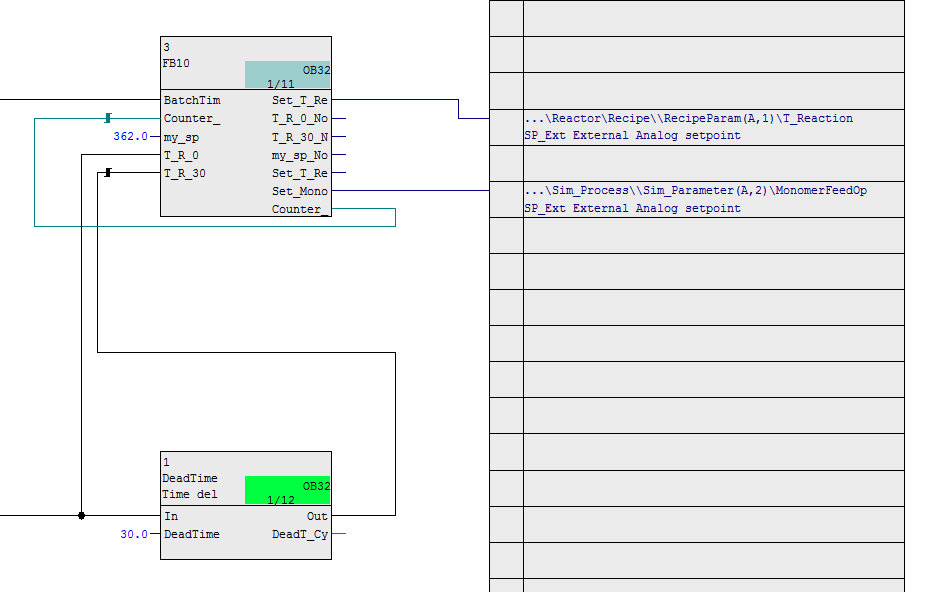}
	\caption[CFC]{
		CFC diagram with the building block (FB10) containing the GPRL policy
	}
	\label{figure:cfc}
\end{figure}

Note that for applying the function to a real production plant, some safety logic and switching between different modes like manual and automatic would be required. 
Integration of this is straight forward and neglected for the following evaluation using the simulation model.

\subsection{Evaluation}
\label{section:evaluation}

To evaluate and visualize the performance of the policy, we tested it on four exemplary intended setpoints, i.e., 352, 358, 362, 365 K.
The intended setpoint $S$ is plotted in red.
The action of the policy $\hat{T}$ is plotted in blue.
The resulting reactor temperature $T$, using the default PID regulator $\hat{T}=S$, is plotted in orange.
The resulting reactor temperature $T$, using the new RL policy given in Eq.~\eqref{eq:policy}, is plotted in green.

Fig.~\ref{figure:eval_352} depicts the results for $S=352$~K.
In the beginning the RL policy changes the temperature setpoint to the maximum value of 365~K, to speed up the heating process of the reactor.
After approximately 80~s, it reduces the setpoint to the minimum value of 352~K, to keep the reactor temperature as close as possible to the intended setpoint.
However, since the setpoint can not be lowered below the minimum of 352~K, the reactor is constantly above the intended setpoint.
In this example, the new RL policy could slightly reduce control deviation by 0.3~\%.
\begin{figure}
	\centering
	\includegraphics[width=3.3in]{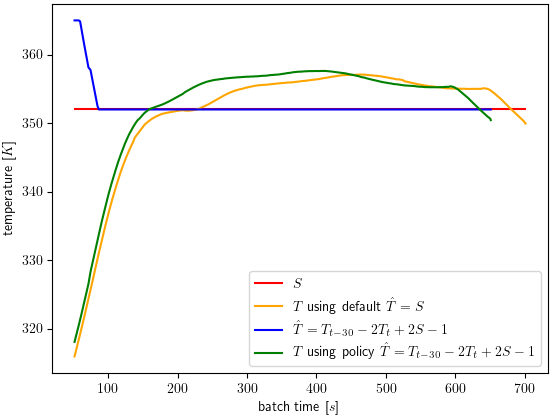}
	\caption[352]{
		Setpoint controls and resulting trajectories for default and GPRL policies. Results are shown for an intended reactor setpoint of 352 K. 
	}
	\label{figure:eval_352}
\end{figure}

A real improvement can be seen for intended setpoint $S=358$~K.
Fig.~\ref{figure:eval_358} shows that the default controller produces huge overshooting, whereas the new RL policy stays close to the intended setpoint over the whole batch time.
Again, the RL policy is increasing the setpoint to the maximum at the beginning just to reduce it to the minimum after 100~s.
After 400~s we can see that the RL policy is starting to slightly increase the setpoint again to counteract the decrease in reactor temperature at the end of the batch.
The control deviation can be reduced by 33.5~\%.
\begin{figure}
	\centering
	\includegraphics[width=3.3in]{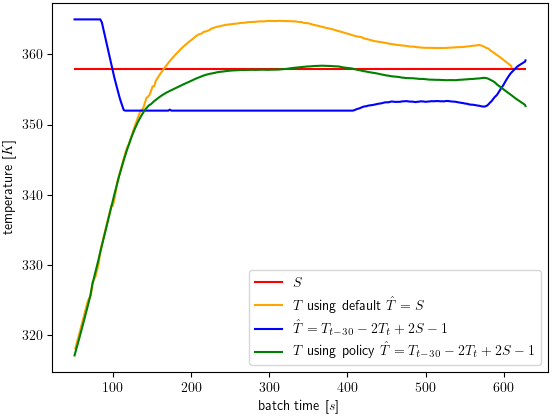}
	\caption[358]{
		Setpoint controls and resulting trajectories for default and GPRL policies. Results are shown for an intended reactor setpoint of 358 K. 
	}
	\label{figure:eval_358}
\end{figure}

In Fig.~\ref{figure:eval_362}, we see a similar behavior for intended setpoint $S=362$~K.
The new RL policy changes the setpoint dynamically over the whole batch time to reduce control deviation by 37.7~\%.
\begin{figure}
	\centering
	\includegraphics[width=3.3in]{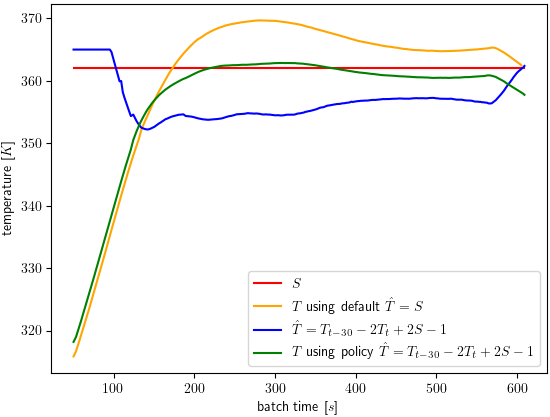}
	\caption[362]{
		Setpoint controls and resulting trajectories for default and GPRL policies. Results are shown for an intended reactor setpoint of 362 K. 
	}
	\label{figure:eval_362}
\end{figure}

The final example is intended setpoint $S=365$~K in Fig.~\ref{figure:eval_365}.
Here, the new RL policy reduces control deviation by 30.1~\%.
Note that in this example, it is very important not to overshoot too much, since depending on the product in the reactor, too high temperatures can spoil the process and ruin the quality of the final product.
\begin{figure}
	\centering
	\includegraphics[width=3.3in]{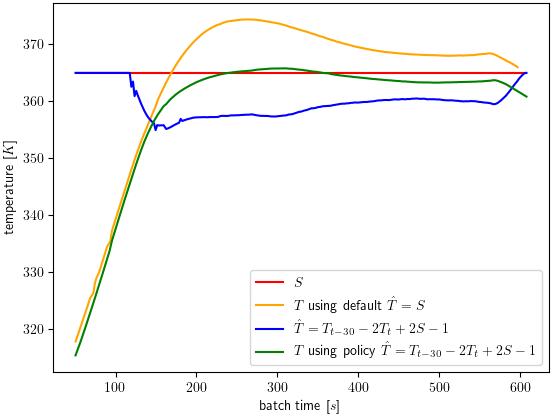}
	\caption[365]{
		Setpoint controls and resulting trajectories for default and GPRL policies. Results are shown for an intended reactor setpoint of 365 K. 
	}
	\label{figure:eval_365}
\end{figure}

To conclude, we can say that the control deviation has been reduced for all intended setpoints tested.
The policy was automatically learned from safely generated batch data, easy to implement, and reduced control deviation by up to 37~\%.
	
	\section{Conclusion}

The experiments using GPRL on a SIMATIC PCS 7 polymerization reactor template show that it is possible to generate human-interpretable policies from existing process data fully automatically.
In contrast to widely-known RL methods, which produce black-box value functions and/or policies, with GPRL, the domain experts can stay in the optimization loop, since the results are convenient to validate and implement.
Utilizing such trustworthy AI methods will help bring state-of-the-art ML algorithms into real-world industry applications, to leverage the optimization potentials which are to be expected in many domains.
	
	\section*{Acknowledgments}
	The project this report is based on was supported with funds from the German Federal Ministry of Education and Research under project number 01IS18049A. 
	The sole responsibility for the report's contents lies with the authors.
	
	\bibliography{bibliography}
	
\end{document}